\pdfoutput=1
\documentclass[11pt]{article}

\usepackage[preprint]{acl}

\usepackage{times}
\usepackage{latexsym}

\usepackage[T1]{fontenc}
\usepackage[utf8]{inputenc}

\usepackage{microtype}

\usepackage{inconsolata}

\usepackage{graphicx}

\usepackage{amsmath}
\usepackage{algorithm}
\usepackage{algpseudocode}
\usepackage{booktabs}
\usepackage[table,xcdraw]{xcolor}
\usepackage{listings}
\usepackage[normalem]{ulem}
\useunder{\uline}{\ul}{}
\usepackage{enumitem}

\title{Tree-of-Text: A Tree-based Prompting Framework for Table-to-Text Generation in the Sports Domain}

\author{
 \textbf{Shang-Hsuan Chiang\textsuperscript{1}},
 \textbf{Tsan-Tsung Yang\textsuperscript{1}},
 \textbf{An-Zi Yen\textsuperscript{1}},
 \textbf{Wen-Chih Peng\textsuperscript{1}},
\\
 \textsuperscript{1}National Yang Ming Chiao Tung University,
\\
 \small{
   \textbf{Correspondence:} \href{mailto:andy10801@gmail.com}{andy10801@gmail.com}
 }
}

\begin{document}
\maketitle

\begin{abstract}

Generating sports game reports from structured tables is a complex table-to-text task that demands both precise data interpretation and fluent narrative generation. Traditional model-based approaches require large, annotated datasets, while prompt-based methods using large language models (LLMs) often struggle with hallucination due to weak table comprehension. To overcome these challenges, we propose Tree-of-Text, a tree-structured prompting framework that guides LLMs through a three-stage generation process: (1) Content Planning, where relevant operations and arguments are selected from the input tables; (2) Operation Execution, which breaks down large tables into manageable sub-tables; and (3) Content Generation, where short textual outputs are merged and rewritten into a cohesive report. Experiments show that our method outperforms existing methods on ShuttleSet+, leads in RG and CO metrics on RotoWire-FG, and excels in CS and CO on MLB with roughly 40\% of the time and cost of Chain-of-Table. These results demonstrate the effectiveness and efficiency of Tree-of-Text and suggest a promising direction for prompt-based table-to-text generation in the sports domain.
\end{abstract}

\section{Introduction}

% 表格是一種廣泛使用的資料格式，其透過 rows and columns 來儲存資訊的方法很有效。然而，相較於文字，人類很難直接從表格中理解資訊。因此，table-to-text generation 是個相當重要的任務，將結構化的表格轉成非結構化的文字，並且可以應用在多個領域上，像是 Sports Game Report、金融財報、醫療分析等等。

% Tables are a widely used data format \citep{cafarella2008webtables}, efficiently storing information through rows and columns. 
% However, compared to text, humans find it more challenging to directly interpret information from tables.
% Therefore, table-to-text generation is a crucial task that converts structured tables into unstructured text, enhancing human readability and comprehension.

% table-to-text generation is a crucial task that converts structured tables into unstructured text, enhancing human readability and comprehension. 在 table-to-text generation 的資料集中，我們特別聚焦在 sports domain 上，像是 RotoWire-FG, MLB, and ShuttleSet+。這些資料集具有較長的文字輸出以及較高的資料保真性等特性，讓這個任務變得更加有挑戰。圖 \ref{} 展示了一個 ShuttleSet+ 的範例，包含多個表格資料以及一個對應這場比賽的人工報導。

Writing sports game reports requires journalists to analyze match data and craft engaging reports under tight deadlines. 
Beyond conveying scores and player performance, they must construct compelling narratives that highlight key moments.
Automating this process could greatly improve the efficiency and accessibility of sports journalism.
However, converting structured match data into natural language remains challenging. 
Sports reporting also demands adherence to journalistic conventions, integrating game flow, player dynamics, and contextual insights, which require reasoning and advanced text organization skills.

\begin{figure*}[t]
  \includegraphics[width=\linewidth]{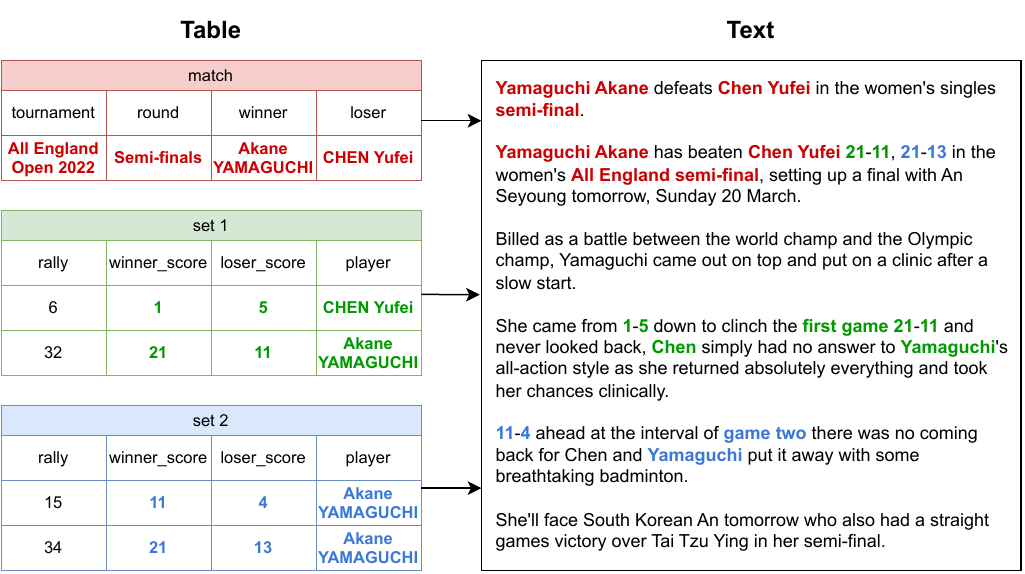}
  \caption{An example from ShuttleSet+, which includes multiple structured tables containing match data along with a corresponding human-written report for the game.}
  \label{fig:example}
\end{figure*}

Thus, sports game report generation is a complex table-to-text generation task involving not only data transformation but also discourse structuring, content selection, and information organization. 
Effectively generating sports articles requires balancing structured data processing with the storytelling aspects of journalism to ensure accuracy and readability.
% table-to-text generation is a crucial task that converts structured tables into unstructured text, enhancing human readability and comprehension.
In this study, we focus specifically on the sports domain, utilizing datasets such as ShuttleSet+, RotoWire-FG \citep{wiseman-etal-2017-challenges}, and MLB \citep{puduppully-etal-2019-data}.
These datasets are characterized by high data fidelity and longer textual outputs, making the task more challenging.
Figure \ref{fig:example} presents an example from ShuttleSet+.
The text contained in the tables is highlighted in \textbf{bold}, with different colors used to distinguish information from different tables.

For this task, numerous model-based methods have been proposed, such as NCP \citep{puduppully2019data}, NDP \citep{CHEN2021106610}, DUV \citep{gong-etal-2020-enhancing}, Macro \citep{10.1162/tacl_a_00381}, and SeqPlan \citep{puduppully-etal-2022-data}.
However, these approaches require large amounts of training data, making them impractical when datasets are scarce and costly to collect.
With advancements in large language models (LLMs), prompt-based methods have become increasingly popular, including Zero-shot, One-shot, and Few-shot learning \citep{brown2020language}, as well as techniques like Chain-of-Thought \citep{NEURIPS2022_9d560961}, Tree-of-Thought \citep{NEURIPS2023_271db992}, and Chain-of-Table \citep{wang2024chainoftable}.
Although these methods are widely used, in the task of generating sports game reports, LLMs fail to effectively analyze and comprehend information within tables, resulting in the generation of text that is inaccurate or not present in the tables, commonly known as the hallucination issue.

% 為了解決這些問題，我們提出了一個新的方法 Tree-of-Text，並將任務分成三個階段：Content Planning, Operation Execution, and Content Generating。首先，在 Content Planning 階段，LLM 會根據表格規劃子節點的 operation。接著，在 Operation Execution 階段，會依序執行規劃的 operation 更新表格，並分別傳給子節點。最後，在 Content Generating 階段，LLM 會根據表格生成文字，並回傳給父節點，父節點再使用 LLM 將回傳的文字合併並重寫成新的文字。

To address these challenges, we propose Tree-of-Text, a novel framework inspired by the "divide and conquer" concept of merge sort \citep{10.1145/355602.361317}, which divides the task into three stages: Content Planning, Operation Execution, and Content Generating.
First, in the Content Planning stage, the LLM plans the operations for child nodes based on the table structure.
Second, in the Operation Execution stage, these selected operations are executed respectively to update the table, which is then passed to the child nodes.
If the \texttt{write()} operation is used or the maximum depth is reached, the LLM generates text based on the table and returns it to the parent node.
Finally, in the Content Generating stage, the LLM merges and rewrites these texts into a new text, which is then returned to the parent node, continuing recursively until reaching the root node.
The final output is the text returned from the root node.

% 因為最後會建立出一個樹狀的結構，所以我們將這個方法命名為 Tree-of-Text。

% Since this approach ultimately forms a tree structure, we name it Tree-of-Text.

% Tree-of-Text 很好的利用了樹狀的結構，將大表格拆分成小表格，增加 LLM 的表格理解能力。
% 此外，我們利用 merge and rewrite 文字的方式，產生更長且內容完整的文字。
% 在 RotoWire-FG, MLB, and ShuttleSet+ 的實驗中，Tree-of-Text 勝過了其他 prompt-based baseline，達到 state of the art 的表現。
% 透過一些優化，我們的方法在 time and cost 還比 Tree-of-Thought 和 Chain-of-Table 更低。證明了我們的方法在效能以及效率上的優勢。

Tree-of-Text effectively leverages a hierarchical tree structure to decompose large tables into smaller sub-tables, enhancing the LLM's ability to comprehend tabular information.
Additionally, we employ a merge-and-rewrite approach to generate longer and more comprehensive reports.
Experimental results demonstrate that Tree-of-Text outperforms other prompt-based baselines on the ShuttleSet+, RotoWire-FG, and MLB datasets. 
Furthermore, with optimizations, our method achieves lower time and cost compared to Tree-of-Thought and Chain-of-Table, highlighting its advantages in both effectiveness and efficiency.

% 總結，我們統整了這篇論文的三個主要貢獻：
% 1. 在 table-to-text generation for sports game report 的任務上，我們提出了 Tree-of-Text，有效的解決了現有的問題。
% 2. 我們提出了一個新的資料集 ShuttleSet+，源自 ShuttleSet22 \citep{ShuttleSet22}，包含了 58 場羽球比賽的 rally-level data 以及對應的人工報導。
% 3. Tree-of-Text outperforms other prompt-based baselines on RotoWire-FG, MLB, and ShuttleSet+ datasets, with relatively lower time and cost. 證明了我們的方法在效能以及效率上的優勢。

We summarize the three main contributions of this paper:

\begin{itemize}[nolistsep]
    \item In the task of table-to-text generation in the sports domain, we introduce Tree-of-Text, a novel framework that recursively decomposes tables into smaller sub-tables, generates short textual descriptions for each sub-table, and merges these short texts into a complete report.
    \item We introduce a new sports domain dataset, ShuttleSet+, containing rally-level data from 58 badminton matches along with the corresponding human-written reports.
    \item Tree-of-Text outperforms other prompt-based baselines on the ShuttleSet+, RotoWire-FG, and MLB datasets while maintaining relatively lower time and cost, demonstrating its superiority in both effectiveness and efficiency.
\end{itemize}

\section{Related Work}

\subsection{Table-to-Text Generation}

% table-to-text generation 任務的目標將結構化的表格轉換成非結構化的文字，通常會分成兩個階段：Content Planning 和 Content Generation \citep{10215344}。Content Planning 也就是 What to say, means to analyze and filter given structured data, from which all or part of the data is selected for abstraction and association. Content Generation 也就是 How to say, refers to accurately and fluently describe the selected data through natural language. Tree-of-Text 也參考這個原則來設計架構。

The goal of the table-to-text generation task is to convert structured tables into unstructured text, typically following a two-stage process: Content Planning and Content Generating \citep{10215344}. 
Content Planning, or ``What to say,'' involves analyzing and filtering the given structured data, selecting relevant information for abstraction and association. 
Content Generating, or ``How to say,'' focuses on accurately and fluently describing the selected data using natural language. 
% Tree-of-Text follows this principle in its architectural design.

% table-to-text generation 的資料集有很多。像是 WikiBio \citep{lebret2016neural} is made up of biographical tables along with the first sentence of each biography article from Wikipedia. ToTTo \citep{parikh2020totto} includes tables with multiple patterns, and completes the text generation innovation of controlled cells in terms of content selection. TabFact \citep{Chen2020TabFact}, each table corresponds to two hypotheses for studying fact verification.

% There are numerous datasets available for table-to-text generation, such as WikiBio~\citep{lebret2016neural}, ToTTo~\citep{parikh2020totto}, and TabFact~\citep{Chen2020TabFact:}.
% 不過我們這篇論文更聚焦在 domain-specific 的資料集上，特別是 for sports game report。
% For example，RotoWire-FG \citep{wiseman-etal-2017-challenges} is a dataset comprised of human-written NBA basketball game summaries paired with their corresponding box and line scores.
% MLB \citep{puduppully-etal-2019-data} is a dataset that contains baseball statistics along with human-written summaries from the ESPN website.
% ShuttleSet+ is a dataset containing rally-level data from 58 badminton matches along with the corresponding human-written reports.
% 這幾個資料集有兩個共同的特性：高資料保真性和長文字輸出。
% 高資料保真性可以用提供正確且重要的資訊，讓讀者對這場運動比賽有更深入的認識。
% 長文字輸出則可以用更豐富且生動的描述，讓讀者對這場運動比賽有更多的興趣。

There are numerous datasets available for table-to-text generation, such as WikiBio~\citep{lebret2016neural}, ToTTo~\citep{parikh2020totto}, and TabFact~\citep{Chen2020TabFact:}. 
Nevertheless, this paper focuses on domain-specific datasets, particularly for the sports domain. 
For instance, ShuttleSet+, derived from ShuttleSet22 \citep{ShuttleSet22}, contains rally-level data from 58 badminton matches along with corresponding human-written reports.
RotoWire-FG \citep{wiseman-etal-2017-challenges} is a dataset consisting of human-written summaries of NBA basketball games paired with their corresponding box and line scores. 
MLB \citep{puduppully-etal-2019-data} provides baseball statistics accompanied by human-authored summaries from the ESPN website. 
These datasets share two common characteristics: high data fidelity and long textual outputs. 
High data fidelity ensures accurate and important information, enabling readers to gain a deeper understanding of the sports matches.
Long textual outputs, on the other hand, provide rich and vivid descriptions, enhancing reader engagement and interest in the sports games.

\subsection{Model-based Methods}

Although several model-based methods have been proposed in prior work, such as NCP~\citep{puduppully2019data}, NDP~\citep{CHEN2021106610}, DUV~\citep{gong-etal-2020-enhancing}, Macro~\citep{10.1162/tacl_a_00381}, and SeqPlan~\citep{puduppully-etal-2022-data}, our experimental setting involves a limited amount of data (e.g., ShuttleSet+ contains only 58 samples). This makes model training highly challenging and causes model-based approaches to underperform compared to expectations. Therefore, we do not include model-based methods as baselines and instead compare our approach with prompt-based methods.

\subsection{Prompt-based Methods}

% 隨著 LLM 的興起，prompt-based method 也開始越來越多。這篇論文 \citep{brown2020language} 第一次提出了 Zero-shot, One-shot, Few-shot 的方法，只要提供少量的參考範例，LLM 就可以在多種任務上達到不錯的結果。Chain-of-Thought \citep{NEURIPS2022_9d560961} 則是在 prompt 中加入少量的 a series of intermediate reasoning steps，就可以提高 LLM 的推理能力。Tree-of-Thought \citep{NEURIPS2023_271db992} allows LLM to perform deliberate decision making by considering multiple different reasoning paths and self-evaluating choices to decide the next course of action, as well as looking ahead or backtracking when necessary to make global choices. Chain-of-Table \citep{wang2024chainoftable} guides LLMs to iteratively generate operations, updating the table to form a tabular reasoning chain. This allows LLMs to dynamically plan the next operation based on previous results, continuously evolving the table to illustrate the reasoning process.

With the rise of LLMs, prompt-based methods have gained increasing attention. 
\citet{brown2020language} first introduced the Zero-shot, One-shot, and Few-shot approaches, demonstrating that providing some reference examples enables LLMs to achieve strong performance across various tasks. 
Chain-of-Thought~\citep{NEURIPS2022_9d560961} enhances LLM reasoning by incorporating a series of intermediate reasoning steps within the prompt. 
Tree-of-Thought~\citep{NEURIPS2023_271db992} enables LLMs to make deliberate decisions by exploring multiple reasoning paths, self-evaluating choices, and backtracking when necessary to optimize global decision-making. 
Chain-of-Table~\citep{wang2024chainoftable} guides LLMs to iteratively generate operations, updating the table to form a tabular reasoning chain, allowing for dynamic operation planning based on previous results.

% 然而，先前的方法都是直接輸入表格，讓 LLM 難以理解表格的結構，導致了幻覺的問題，不滿足 high data fidelity 的特性。
% 此外，先前的方法都是直接輸出文字，讓 LLM 難以理解全面的資訊，導致了生成的文字較簡短，不滿足 long textual output 的特性。
% 因此，我們提出了 Tree-of-Text 來解決這些問題。

However, previous methods directly input the entire table into the LLM, making it difficult for the model to fully understand the table structure, thereby resulting in hallucination and failing to ensure high data fidelity.
Similarly, these methods directly output the final text in a single step, limiting the model’s ability to process comprehensive information, thus failing to produce sufficiently long and detailed textual outputs.
Therefore, we propose Tree-of-Text to address these challenges.

\section{Tree-of-Text}

\begin{figure*}[t]
  \includegraphics[width=\linewidth]{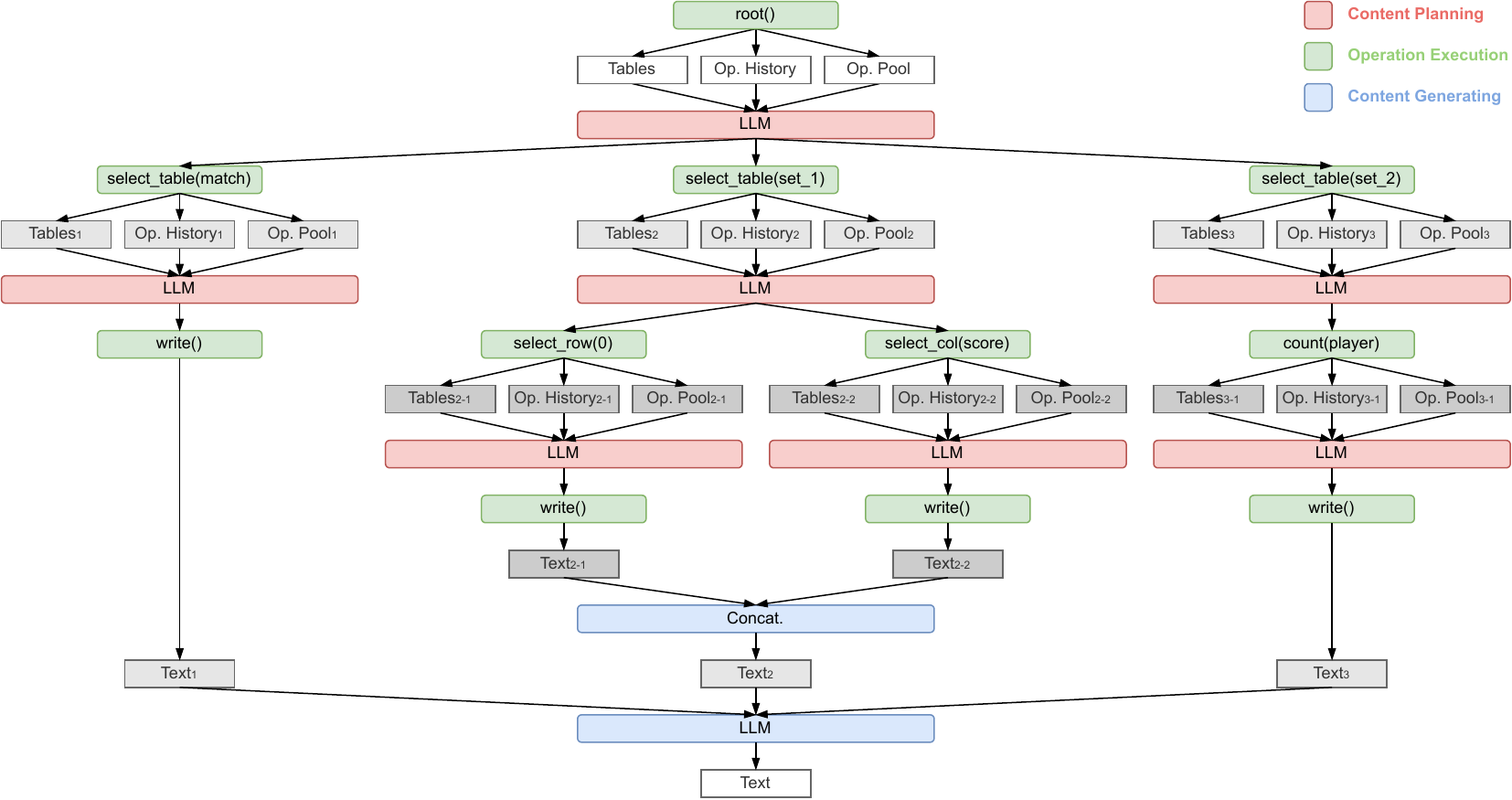}
  \caption{The overall framework of Tree-of-Text, which constructs a tree structure and divides the task into three stages: Content Planning, Operation Execution, and Content Generating. For simplicity, a basic tree structure is shown as an example; in practice, the tree is more complex.}
  \label{fig:overview}
\end{figure*}

% Tree-of-Report 的整體流程圖，我們參考 merge sort "divide and conquer" 的概念，將 Table-to-Text Generation 的任務分成三個階段：Content Planning, Operation Execution, and Content Generating。為了簡化圖片，我們只用一個簡單的樹狀結構來當作例子，實際上會比這個更加複雜。

\begin{figure*}[t]
  \includegraphics[width=\linewidth]{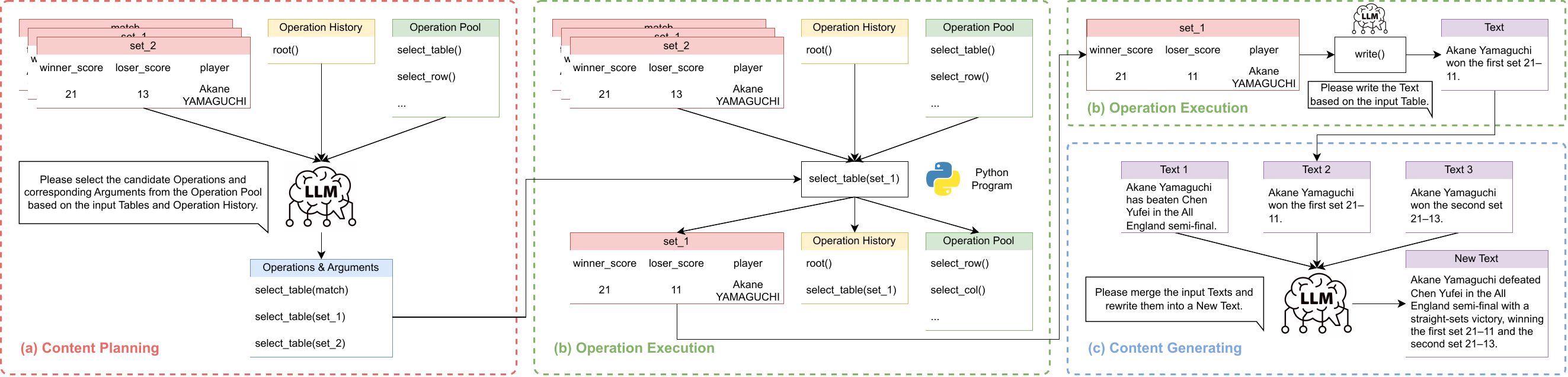}
  \caption{The detailed workflow of Content Planning, Operation Execution, and Content Generating.}
  \label{fig:workflow}
\end{figure*}

% Content Planning, Operation Execution, and Content Generating 的細節流程圖

\subsection{Overview}

In the task of table-to-text generation, the input consists of multiple tables \( T \), and the output is a text \( t \) to describe these tables. 
We propose a method called Tree-of-Text, inspired by the "divide and conquer" concept of merge sort \citep{10.1145/355602.361317}, which divides the task into three stages: Content Planning, Operation Execution, and Content Generating.

In the Content Planning stage, the LLM determines the operations and arguments \( OA \) for the child nodes based on the input tables \( T \), the operation history \( OH \), the operation pool \( OP \), and the depth \( D \). 
The number of child nodes must not exceed the maximum degree \( MAX\_DEGREE \).

In the Operation Execution stage, the operations in \( OA \) are executed respectively to update \( T_i' \), \( OH_i' \), \( OP_i' \), and \( D_i' \), where $i = 1, 2, \ldots, d$, which are then passed to the child nodes. 
This process continues recursively until the depth \( D_i' \) reaches the maximum depth \( MAX\_DEPTH \) or a \texttt{write()} operation is encountered. 
At this point, the LLM writes a short text \( t_i' \) to describe the updated table \( T_i' \), and then returns it to the parent node.

In the Content Generating stage, the LLM merges these texts \( t_i' \) from the child nodes and rewrites them into a new text \( t \). 
The new text \( t \) is then returned to the parent node, and this process continues recursively until returning to the root node. 
The final output is the text returned by the root node.

The overall framework of Tree-of-Text is illustrated in Figure \ref{fig:overview}, and the algorithm is presented in Algorithm \ref{alg:overview}.

\subsection{Content Planning}

Starting from the root node, the inputs consist of the initial tables \( T \gets ( T^j \mid j = 1, 2, \ldots, n ) \), the operation history \( OH \gets ( op \mid op = \text{root()} ) \), the operation pool \( OP \gets ( op \mid op \in \text{operations},\ op \neq \text{root()} ) \), and the depth \( D \gets 0 \). 
Based on these inputs, the LLM determines the operations and arguments for the child nodes, denoted as \( OA \gets ( O_i ( A_i ) \mid O_i \in \text{OP}, i = 1, 2, \ldots, d ) \), where \( d \) represents the degree of this node and must not exceed the maximum degree \( MAX\_DEGREE \). 
The detailed workflow of Content Planning is shown in Figure \ref{fig:workflow} (a), and the example prompt for Content Planning is provided in Appendix \ref{sec:content_planning_prompt}.

\subsection{Operation Execution}

To perform operations that decompose large tables into smaller sub-tables or generate text, we utilize the four operations (select\_row(), select\_col(), count(), sort()) from~\citet{wang2024chainoftable} and introduce four additional ones (root(), select\_table(), filter(), write()). This results in a total of eight operations, defined as follows:

\begin{itemize}[nolistsep]
    \item \texttt{root()}: Does nothing; represents the root node of the tree.
    \item \texttt{select\_table()}: Selects a table by its table name.
    \item \texttt{select\_row()}: Selects rows based on their row indices.
    \item \texttt{select\_col()}: Selects columns based on their column names.
    \item \texttt{count()}: Counts the number of unique values in the specified columns of the tables.
    \item \texttt{sort()}: Sorts rows based on the specified column names and sorting orders.
    \item \texttt{filter()}: Filters rows based on column names, comparison symbols, and values.
    \item \texttt{write()}: Writes a short text based on the tables using the LLM; represents the leaf node of the tree.
\end{itemize}

The operations in \( OA \) are then executed by the Python program respectively to update \( T \), \( OH \), \( OP \), and \( D \), where \( T_i' \gets O_i ( T, A_i ) \), \( OH_i' \gets OH + O_i ( A_i ) \), \( OP_i' \gets OP - O_i() \), \( D_i' \gets D + 1 \).  
The updated \( T_i' \), \( OH_i' \), \( OP_i' \), and \( D_i' \) are then passed to the child nodes, where the steps of Content Planning and Operation Execution are performed again.
This process continues recursively until either the depth \( D \) reaches the maximum depth \( MAX\_DEPTH \) or a \texttt{write()} operation is encountered.
If the depth \( D_i' \) reaches the maximum depth \( MAX\_DEPTH \), the LLM is used to generate a textual description \( t \) of the input table \( T \), which is then returned to the parent node.
Otherwise, if a \texttt{write()} operation is encountered, the LLM writes a short text \( t_i' \) based on the input table \( T \) as well.
Since other child nodes also return texts \( t_i' \), we collect them into a sequence \( t' = ( t_i' \mid i = 1, 2, \ldots, d ) \).
The detailed workflow of Operation Execution is shown in Figure \ref{fig:workflow} (b), and the example prompt for the \texttt{write()} operation is provided in Appendix \ref{sec:write_prompt}.

\subsection{Content Generating}

% The LLM then merges and rewrites these short texts \( t' \) into a new text \( t \), which is passed to the parent node. 並且再次進行 Content Generating
The LLM then merges and rewrites these short texts \( t' \) into a new text \( t \), which is subsequently returned to the parent node for another round of Content Generation.
This recursive process continues until it returns to the root node. 
The text \( t \) returned from the root node is the final output. 
The detailed workflow of Content Generating is shown in Figure \ref{fig:workflow} (c), and the example prompt for Content Generating is provided in Appendix \ref{sec:content_generating_prompt}.

For efficiency considerations, we implemented additional optimizations.  
First, unlike Chain-of-Table, which generates operations first and then arguments, our method generates operations and arguments in one step.  
Second, if a node has only one child node, there is no need to use the LLM for merging. 
Finally, we experimented with an approach where the LLM is used for merging only at the root node, while other nodes simply concatenate texts.  
With these optimizations, Tree-of-Text significantly reduces both time and cost.

\section{Experiment}

\subsection{Dataset}

\subsubsection{ShuttleSet+}

We introduce a new dataset, ShuttleSet+, derived from ShuttleSet22 \citep{ShuttleSet22}. 
ShuttleSet22 is a human-annotated, stroke-level singles dataset for badminton tactical analysis, comprising 140 sets, 3,992 rallies, and 33,612 strokes from 58 matches played between 2018 and 2022. 
Since ShuttleSet22 does not include corresponding textual reports for each match, we collected human-written reports in English for each game from online sources such as the BWF and Olympics websites, and renamed the dataset as ShuttleSet+. 
Compared to RotoWire-FG and MLB, ShuttleSet+ has fewer data samples, representing a low-resource scenario. 
Finally, the dataset is randomly partitioned into training, validation, and test sets with a ratio of 40:9:9.
Further details on data preprocessing for ShuttleSet+ can be found in Appendix~\ref{sec:data_preprocessing_for_shuttleset+}.

\subsubsection{RotoWire-FG}

% RotoWire-FG \citep{wang-2019-revisiting} 是一個源自於 RotoWire \citep{wiseman-etal-2017-challenges} 的資料集，他們 crawl the game summaries from the RotoWire Game Recaps between years 2017-19 and align the summaries with the official NBA boxscore tables. 總共有 7635 筆資料，並隨機分成 train 5340 筆，valid 1147 筆，test 1148 筆。此外，他們也 purify and normalize contents for dataset purification。Data preprocessing for RotoWire-FG is provided in Appendix~\ref{sec:data_preprocessing_for_rotowire} for further details.

The RotoWire-FG dataset \citep{wang-2019-revisiting} is an extension of the original RotoWire dataset \citep{wiseman-etal-2017-challenges}, constructed using basketball game summaries from RotoWire Game Recaps covering the years 2017–2019, and aligned with official NBA box score tables.
The dataset is randomly divided into 5,340 for training, 1,147 for validation, and 1,148 for testing.
Compared to RotoWire, RotoWire-FG includes an additional 2.6K games and incorporates line-score replenishment to enhance data completeness.
Moreover, the content is further purified and normalized to improve data quality.
Further details on data preprocessing for RotoWire-FG can be found in Appendix~\ref{sec: data_preprocessing_for_rotowire}.

\subsubsection{MLB}

The MLB dataset \citep{puduppully-etal-2019-data} contains baseball statistics paired with human-written summaries in English sourced from the ESPN website. 
Compared to RotoWire-FG, it is approximately three times larger, featuring a broader vocabulary and longer summaries. 
The dataset is divided into 22,821 training, 1,739 validation, and 1,744 testing instances.
Further details on data preprocessing for MLB can be found in Appendix~\ref{sec:data_preprocessing_for_mlb}.

\subsection{Evaluation Metric}

\subsubsection{Automatic Evaluation}

To automatically evaluate the informational alignment between two texts, we adopt the Information Extraction (IE) metrics proposed by \citet{wiseman-etal-2017-challenges}.
These metrics are based on the output of an IE model, which extracts information pairs, formatted as \texttt{(table|column|value)}, from the generated text.

% 因為我們沒有足夠的資料去訓練一個新的 IE model for ShuttleSet+，所以我們提出使用 LLM 來當作 IE model 的替代方案。

Due to the lack of sufficient data to train a new IE model for ShuttleSet+, we propose an alternative approach that leverages the LLM as a substitute for the IE model.
To validate its reliability, we manually annotated a set of information and compared it with that extracted by the LLM, finding that it achieved over 70\% on all evaluation metrics with few-shot prompting. 
Based on this experiment, we consider that using an LLM as an IE model is a reliable alternative.
The full experimental results are provided in Appendix~\ref{sec:LLM_IE_model}.

In the experiment, let \( \hat{t} \) denote the generated text and \( t \) symbolize the gold text. 
Relation Generation (RG) evaluates the count (\#) and precision (P\%) of information extracted from \( \hat{t} \) that is present in the input table \( T \), representing the amount and accuracy of information in the generated text. 
Content Selection (CS) assesses the precision (P\%), recall (R\%), and F1 score (F\%) of information extracted from \( \hat{t} \) that also appears in \( t \), indicating the information similarity between the generated text and the gold text. 
Content Ordering (CO) quantifies the complement of the Damerau-Levenshtein Distance (DLD\%) \citep{damerau1964} between information extracted from \( \hat{t} \) and \( t \), meaning the ordering similarity between the generated text and the gold text. 
Higher values of RG, CS, and CO indicate better effectiveness.

Additionally, to evaluate the efficiency of each method, we compute the average time (in seconds) and cost (in \$0.001 USD) required to generate a text. 
The time refers to the end-to-end duration from the table input to the generated text.
The cost is estimated based on the API Pricing published by OpenAI \citep{openai_api_pricing}.
Lower time and cost values mean better efficiency.

\subsubsection{Human Evaluation}

% 為了進一步驗證我們方法的有效性，我們請了三位測試者來進行 human evaluation。三位測試者都精通英文，並都具有大學以上的學歷，我們會先給予測試者明確的指示，讓他們清楚實驗細節，並且先進行一個小測驗，確保測試者符合實驗資格。此外，我們給予了測試者高於當地最低薪資的酬勞，以確保勞工權益。
% human evaluation 實驗分成兩個部份。首先，我們從三個資料集中隨機挑選十場比賽，然後挑選每場比賽的 gold text 以及用 Chain-of-Thought、Tree-of-Thought、Chain-of-Table 和 Tree-of-Text 生成的文字並隨機打散。第一部份，我們請使用者根據表格去計算每個文字的 supported facts (consistent with the table) and contradicted facts (inconsistent with the table)，然後我們將分數取平均。第二部份，我們請使用者從五個文字中選出最好和最差的文字，根據 three aspects: Coherence (how logically and smoothly the ideas and events are connected throughout the report), Conciseness (how effectively a report conveys information using as few words as necessary, without unnecessary repetition or irrelevant details), and Grammaticality (whether the text follows the rules of standard English grammar)，接著使用 Best-Worst scaling \cite{Louviere_Flynn_Marley_2015} 轉換成 +100 到 -100 的分數，越高越好。

To further validate the effectiveness of our proposed method, we conducted a human evaluation study involving three annotators. All annotators are fluent in English and possess at least a university-level education. Before the evaluation, we provided detailed instructions outlining the task procedure and conducted a preliminary qualification test to ensure that participants fully understood the experimental protocol. Additionally, we compensated all annotators at a rate above the local minimum wage to ensure fair labor conditions.

% 我們的 human evaluation 參考了 \citet{puduppully-etal-2022-data} 的作法，主要分成兩個部份。

Our human evaluation follows the methodology proposed in \citet{puduppully-etal-2022-data} and is divided into two parts. First, we randomly selected ten matches from each of the three datasets. For each match, we compiled one gold reference summary and four generated summaries from Chain-of-Thought, Tree-of-Thought, Chain-of-Table, and Tree-of-Text, then randomly shuffled their order. In the first part of the evaluation, annotators were asked to analyze each summary against the corresponding tables and count the number of Supported Facts (i.e., statements consistent with the table) and Contradicted Facts (i.e., statements inconsistent with the table). We report the average scores across all evaluations. In the second part, annotators were instructed to select the best and worst summary from the five options based on three criteria: Coherence (how logically and smoothly the ideas and events are connected throughout the report), Conciseness (how effectively a report conveys information using as few words as necessary, without unnecessary repetition or irrelevant details), and Grammaticality (whether the text follows the rules of standard English grammar). The results were then converted into a score between +100 and -100 using the Best-Worst Scaling method \cite{Louviere_Flynn_Marley_2015}, with higher scores indicating better quality.

\subsection{Implementation Detail}

For all datasets, Tree-of-Text employs gpt-4o-mini \citep{openai2024gpt4omini} as the backbone large language model. We set the maximum depth to 5 and the maximum degree to 5, utilize the full operation pool, and represent all tables in CSV format. 
% Furthermore, we adapt the prompts used for Content Planning, \texttt{write()}, and Content Generating stages to suit the specific characteristics of each dataset.
All of our experiments are single-run.

\begin{table*}[]
\small

\resizebox{\linewidth}{!}{%
\begin{minipage}{\linewidth}
    \centering
    % \textbf{ShuttleSet+ Results} \\[0.5em]
    \begin{tabular}{@{}ccccccccc@{}}
    \toprule
    \textbf{ShuttleSet+} & \textbf{RG \#} & \textbf{RG P\%} & \textbf{CS P\%} & \textbf{CS R\%} & \textbf{CS F\%} & \textbf{CO DLD\%} & \textbf{Time}  & \textbf{Cost}  \\ \midrule
    Zero-shot            & 14.00          & 82.69           & 65.81           & 75.90           & 69.85           & 48.25             & 7.53           & {\ul 0.86}           \\
    One-shot             & 13.56          & 81.61           & 65.57           & 72.75           & 68.36           & 49.40             & {\ul 6.59}           & 1.12           \\
    Few-shot             & 13.78          & 82.81           & 66.57           & 75.33           & 70.20           & {\ul 53.00}       & \textbf{6.00}           & 2.20           \\
    Chain-of-Thought     & 12.33          & 81.35           & 71.33           & 73.58           & 71.41           & 52.27             & 6.68           & \textbf{0.81}           \\ \midrule
    Tree-of-Thought      & 13.67          & 79.40           & 61.98           & 69.31           & 63.91           & 46.50             & 63.11    & 9.62     \\
    Chain-of-Table       & {\ul 15.89}    & {\ul 94.31}     & {\ul 69.22}     & {\ul 79.74}     & {\ul 73.14}     & 40.42             & 73.67 & 14.44 \\ \midrule
    \rowcolor[HTML]{FFFFC7} 
    Tree-of-Text         & \textbf{16.78} & \textbf{95.79}  & \textbf{73.74}  & \textbf{85.16}  & \textbf{78.03}  & \textbf{69.30}    & 29.04          & 5.71           \\ \bottomrule
    \end{tabular}%
\end{minipage}
}

\vspace{1em} % 在三個表格之間增加垂直間距

\resizebox{\linewidth}{!}{%
\begin{minipage}{\linewidth}
    \centering
    % \textbf{RotoWire Results} \\[0.5em]
    \begin{tabular}{@{}ccccccccc@{}}
    \toprule
    \textbf{RotoWire-FG} & \textbf{RG \#} & \textbf{RG P\%} & \textbf{CS P\%} & \textbf{CS R\%} & \textbf{CS F\%} & \textbf{CO DLD\%} & \textbf{Time}  & \textbf{Cost}  \\ \midrule
    Zero-shot            & {\ul 47.13}    & 94.13           & 63.44           & \textbf{63.02}  & 63.23           & 27.74             & 5.59           & {\ul 0.63}           \\
    One-shot             & 42.90          & 94.24           & 65.78           & 62.73           & 64.22           & 28.48             & \textbf{5.07}           & 0.90           \\
    Few-shot             & 42.94          & 94.48           & 69.22           & 61.12           & \textbf{64.92}  & 29.71             & {\ul 5.38}           & 1.48           \\
    Chain-of-Thought     & \textbf{49.29} & 93.54           & 64.06           & {\ul 62.98}     & 63.51           & 28.51             & 7.76           & \textbf{0.61}           \\ \midrule
    Tree-of-Thought      & 45.73          & 94.55           & 65.82           & 62.80           & {\ul 64.28}     & 28.55             & 54.62    & 8.18     \\
    Chain-of-Table       & 36.94          & {\ul 94.95}     & {\ul 70.93}     & 56.26           & 62.75           & {\ul 30.19}       & 63.75 & 12.54 \\ \midrule
    \rowcolor[HTML]{FFFFC7} 
    Tree-of-Text       & 34.68          & \textbf{95.11}  & \textbf{74.65}  & 55.88           & 63.92           & \textbf{31.90}    & 33.56          & 7.69           \\ \bottomrule
    \end{tabular}%
\end{minipage}
}

\vspace{1em} % 在三個表格之間增加垂直間距

\resizebox{\linewidth}{!}{%
\begin{minipage}{\linewidth}
    \centering
    % \textbf{MLB Results} \\[0.5em]
    \begin{tabular}{@{}ccccccccc@{}}
    \toprule
    \textbf{MLB}     & \textbf{RG \#} & \textbf{RG P\%} & \textbf{CS P\%} & \textbf{CS R\%} & \textbf{CS F\%} & \textbf{CO DLD\%} & \textbf{Time}  & \textbf{Cost}  \\ \midrule
    Zero-shot        & 84.21          & 87.68           & 60.92           & 61.95           & 61.29           & 58.07             & 12.13          & \textbf{1.46}           \\
    One-shot         & 82.80          & 87.25           & 60.68           & 61.04           & 60.86           & 58.28             & 9.98           & 3.05           \\
    Few-shot         & {\ul 84.54}    & 88.18           & 60.43           & 61.36           & 60.75           & 57.25             & \textbf{8.47}           & 4.85           \\
    Chain-of-Thought & \textbf{84.67} & {\ul 88.61}     & 61.93           & 62.57           & 61.98           & 59.21             & {\ul 9.34}           & {\ul 1.73}           \\ \midrule
    Tree-of-Thought  & 80.38          & 86.25           & {\ul 64.33}     & {\ul 63.29}     & {\ul 63.37}     & {\ul 59.64}       & 47.63    & 7.25     \\
    Chain-of-Table   & 80.99          & \textbf{88.67}  & 64.02           & 60.81           & 61.36           & 57.55             & 55.60 & 10.80 \\ \midrule
    \rowcolor[HTML]{FFFFC7} 
    Tree-of-Text     & 80.11          & 87.04           & \textbf{65.84}  & \textbf{63.42}  & \textbf{63.69}  & \textbf{59.83}    & 29.18          & 6.77           \\ \bottomrule
    \end{tabular}%
\end{minipage}
}

\caption{The results of automatic evaluation for ShuttleSet+, RotoWire-FG, and MLB datasets, where the best scores are highlighted in \textbf{bold}, the second-best scores are \underline{underlined}, and our method is marked with a yellow background.}
\label{tab:automatic_evaluation}

\end{table*}

\begin{table*}[]
\small

\resizebox{\linewidth}{!}{%
\begin{minipage}{\linewidth}
    \centering
    % \textbf{ShuttleSet+ Results} \\[0.5em]
    \begin{tabular}{cccccc}
    \toprule
    \textbf{ShuttleSet+} & \textbf{\#Supp.} & \textbf{\#Cont.} & \textbf{Cohe.}   & \textbf{Conc.}   & \textbf{Gram.}   \\ \midrule
    Gold                 & 3.78            & \textbf{0.78}   & \textbf{100.00} & \textbf{100.00} & \textbf{100.00} \\
    Chain-of-Thought     & 3.67            & 2.33            & -100.00         & \textbf{100.00}         & -100.00         \\
    Tree-of-Thought      & 6.56            & 2.33            & -66.67          & {\ul -44.44}         & 0.00            \\
    Chain-of-Table       & {\ul 7.00}      & 2.11            & {\ul 77.78}     & -100.00 & 50.00           \\ \midrule
    \rowcolor[HTML]{FFFFC7} 
    Tree-of-Text         & \textbf{8.22}   & {\ul 1.00}      & \textbf{100.00} & -100.00 & {\ul 55.55}     \\ \bottomrule
    \end{tabular}%
\end{minipage}
}

\vspace{1em} % 在三個表格之間增加垂直間距

\resizebox{\linewidth}{!}{%
\begin{minipage}{\linewidth}
    \centering
    % \textbf{RotoWire Results} \\[0.5em]
    \begin{tabular}{cccccc}
    \toprule
    \textbf{RotoWire-FG} & \textbf{\#Supp.} & \textbf{\#Cont.} & \textbf{Cohe.}   & \textbf{Conc.}   & \textbf{Gram.}   \\ \midrule
    Gold                 & 9.00            & \textbf{0.44}   & \textbf{100.00} & \textbf{100.00} & \textbf{100.00} \\
    Chain-of-Thought     & 10.22           & 2.11            & -100.00         & {\ul 66.67}          & -66.67          \\
    Tree-of-Thought      & \textbf{14.67}  & 1.44            & -50.00          & 50.00          & -50.00          \\
    Chain-of-Table       & 11.11           & {\ul 0.56}      & {\ul 66.67}     & -50.00           & 0.00            \\ \midrule
    \rowcolor[HTML]{FFFFC7} 
    Tree-of-Text         & {\ul 13.33}     & \textbf{0.44}   & \textbf{100.00} & -77.78     & {\ul 55.56}     \\ \bottomrule
    \end{tabular}%
\end{minipage}
}

\vspace{1em} % 在三個表格之間增加垂直間距

\resizebox{\linewidth}{!}{%
\begin{minipage}{\linewidth}
    \centering
    % \textbf{MLB Results} \\[0.5em]
    \begin{tabular}{cccccc}
    \toprule
    \textbf{MLB}     & \textbf{\#Supp.} & \textbf{\#Cont.} & \textbf{Cohe.}   & \textbf{Conc.}   & \textbf{Gram.}   \\ \midrule
    Gold             & 6.67            & \textbf{0.33}   & \textbf{100.00} & \textbf{100.00} & \textbf{100.00} \\
    Chain-of-Thought & \textbf{13.44}  & 1.67            & -100.00         & {\ul 50.00}          & -100.00         \\
    Tree-of-Thought  & {\ul 11.00}     & 1.44            & -66.67          & 0.00          & 0.00            \\
    Chain-of-Table   & 7.89            & {\ul 0.89}      & {\ul 50.00}     & -33.33        & {\ul 44.44}     \\ \midrule
    \rowcolor[HTML]{FFFFC7} 
    Tree-of-Text     & 7.11            & {\ul 0.89}      & \textbf{100.00} & -66.67      & \textbf{100.00} \\ \bottomrule
    \end{tabular}%
\end{minipage}
}

\caption{The results of human evaluation for ShuttleSet+, RotoWire-FG, and MLB datasets, where the best scores are highlighted in \textbf{bold}, the second-best scores are \underline{underlined}, and our method is marked with a yellow background.}
\label{tab:human_evaluation}

\end{table*}

\subsection{Quantitative Result}

\subsubsection{Automatic Evaluation}

We compared Tree-of-Text with other prompt-based methods as baselines on the ShuttleSet+, RotoWire-FG, and MLB. The results of automatic evaluation are shown in Table~\ref{tab:automatic_evaluation}.
We compared Tree-of-Text with other prompt-based methods as baselines on the ShuttleSet+, RotoWire-FG, and MLB datasets. The results of the automatic evaluation are shown in Table~\ref{tab:automatic_evaluation}.
From the automatic evaluation, we observe that Tree-of-Text achieves the best performance across all metrics on the ShuttleSet+ dataset, attains the highest scores in RG and CO on RotoWire-FG, and outperforms in CS and CO on MLB, demonstrating its overall superiority in effectiveness.
This is mainly because the RotoWire-FG and MLB datasets contain larger amounts of data and a variety of article types, so using the same set of configurations can not adapt well to different types of articles, resulting in a lower average score.

% 從 automatic evaluation 中我們可以看到，RG 雖然在 ShuttleSet+ 和 RotoWire-FG 是最高的，但在 MLB 並不是最好的。
% 這是因為 MLB 的文字比較內容較長且包含較多資訊，對於 Tree-of-Text 來說比較有挑戰性。
% This demonstrates the significance of Content Planning, which selects operations to extract relevant information.

Although Tree-of-Text achieves the highest RG scores on ShuttleSet+ and RotoWire-FG, it does not perform best on the MLB dataset.
% This can be attributed to the longer and more information-dense summaries in MLB, which pose greater challenges for Tree-of-Text.
These findings highlight the importance of the Content Planning stage, which selects appropriate operations to extract relevant information from the tables.

% 另外，Tree-of-Text 的 CS 在 ShuttleSet+ 和 MLB 的表現最好，但在 RotoWire-FG 的表現並不好。
% 原因是 Tree-of-Text 的 RG # 比較小，導致 CS P\% 高但 CS R\% 低，因此 CS F\% 也比較低。
% This reveals the significance of Operation Execution, which extracts more detailed information by executing operations to decompose the tables.

Additionally, Tree-of-Text achieves the best CS performance on ShuttleSet+ and MLB, but not on RotoWire-FG. 
% This is primarily because Tree-of-Text generates less information (lower RG \#), resulting in higher CS P\% but lower CS R\%, and consequently a lower CS F\%. 
This highlights the importance of the Operation Execution stage, which extracts more fine-grained information by executing operations to decompose the tables.

As for CO, Tree-of-Text achieves the highest score across all datasets. 
This shows the significance of Content Generating, which merges and rewrites texts to maintain the original structure and order of the tables.

Furthermore, while Tree-of-Text does not have the lowest time and cost, it is still lower than Tree-of-Thought and Chain-of-Table. 
For example, on ShuttleSet+, Tree-of-Text achieves approximately 40\% of Chain-of-Table's time and cost, showing its advantage in efficiency. 
This improvement is attributed to the optimizations that significantly reduce the time and cost of the Tree-of-Text.

\subsubsection{Human Evaluation}

% Table \ref{tab:human_evaluation} shows the results of human evaluation for ShuttleSet+, RotoWire-FG, and MLB datasets. 首先，我們發現 Tree-of-Text 的 supported facts 在 ShuttleSet+ 和 RotoWire-FG 分別是最高和次高，且 contradicted facts 在 RotoWire-FG 最低，在 ShuttleSet+ 和 MLB 則是次低，可見我們的方法可以有效的包含更多資訊的同時，減少 LLM 幻覺的發生。再來，Tree-of-Text 的 Coherence, Conciseness, Grammaticality 在三個資料集上不是最高就是次高，只略遜於 Gold text，再次驗證了我們的方法可以生成更流暢，簡潔且文法正確的文字，接近於人類寫的文字。
% The agreement between raters using Krippendorff’s α was 0.79 for supported facts and contradicting facts, and 0.77 for Coherence, Conciseness, Grammaticality. 表示實驗結果具有可信賴的一致性。

Table~\ref{tab:human_evaluation} presents the human evaluation results on the ShuttleSet+, RotoWire-FG, and MLB datasets. 
First, we observe that Tree-of-Text achieves the highest or second-highest scores in Supported Facts (\#Supp.) on ShuttleSet+ and RotoWire-FG, and the lowest or second-lowest scores in Contradicted Facts (\#Cont.) on ShuttleSet+, RotoWire-FG, and MLB. 
These results suggest that our method effectively includes more factual information while reducing the incidence of LLM hallucinations. 
Furthermore, Tree-of-Text ranks first or second in Coherence (Cohe.) and Grammatically (Gram.) across all three datasets, only slightly behind the gold reference texts. 
However, Tree-of-Text performs poorly in terms of Conciseness (Conc.), mainly because the generated text tends to be longer and more detailed. 
This further confirms that our method generates text that is fluent and grammatically correct, but with more details.
The inter-rater agreement, measured by Krippendorff’s $\alpha$, was 0.79 for supported and contradicted facts, and 0.77 for coherence, conciseness, and grammaticality, indicating that our human evaluation falls within an acceptable range.

\subsection{Qualitative Result}

% 圖 \ref{fig:qualitative_result} 呈現了 qualitative result of human-written, Chain-of-Table and Tree-of-Text. 為了方便比較，我們將文字中的資訊標記為粗體，有包含在表格中的資訊標記成綠色，錯誤或不包含在表格中的資訊則標記成紅色。
% 從 qualitative result 中可以看出 Tree-of-Text 和 Chain-of-Table 生成的文字相比，包含更多且更詳細的資訊（球種的使用頻率），而且更正確（Tree-of-Text 只錯了一個，而 Chain-of-Table 錯了七個）。這再次驗證了 Tree-of-Text 可以滿足 sports game reports 高資料保真性和長文字的特性。

% Figure~\ref{fig:qualitative_result} showcases the qualitative results of Human, Chain-of-Table, and Tree-of-Text outputs. 
% For ease of comparison, we mark the information in the text with \textbf{bold}: green indicates information included in the tables, while red indicates errors or information not found in the tables.  

Figure~\ref{fig:qualitative_result} showcases the qualitative results of Human, Chain-of-Table, and Tree-of-Text outputs. 
For ease of comparison, we mark the information in the text with \textbf{bold}: green indicates information included in the tables, while red indicates errors or information not found in the tables.  

% To showcase a case study of our method, we present the qualitative results in the Appendix. 
From the qualitative analysis, we observe that the text generated by Tree-of-Text is more detailed and richer in content compared to the human-written text. Additionally, we also observe that, compared to Chain-of-Table, Tree-of-Text generates more comprehensive and detailed information (e.g., shot type frequencies) and produces more accurate outputs, with only one error compared to seven errors from Chain-of-Table. This further validates that Tree-of-Text can generate text that meets the characteristics of high data fidelity and long-form output for sports game reports.

\section{Conclusion}

% In this paper, we propose Tree-of-Text, a novel framework for table-to-text generation in sports game reports. 
% Inspired by the "divide and conquer" concept of merge sort \citep{10.1145/355602.361317}, our method divides the generation process into three stages: Content Planning, Operation Execution, and Content Generating. 
% By recursively decomposing large tables into smaller sub-tables and merging short texts into a long text, our approach effectively enhances data fidelity and generates coherent long-form outputs.
% Experimental results demonstrate that Tree-of-Text outperforms other prompt-based baselines on ShuttleSet+, RotoWire-FG, and MLB datasets, highlighting its superiority in effectiveness.
% Furthermore, our method achieves approximately 40\% of Chain-of-Table's time and cost, showing its advantage in efficiency. 
% In summary, Tree-of-Text opens a new path for prompt-based table-to-text generation in sports game reports.

In this paper, we propose Tree-of-Text, a novel framework for table-to-text generation in the sports domain, inspired by the ``divide and conquer'' principle of merge sort \citep{10.1145/355602.361317}. Our approach recursively decomposes large tables into smaller sub-tables and merges short texts to generate coherent long-form outputs. Experiments show that Tree-of-Text achieves the best performance on ShuttleSet+, leads in RG and CO on RotoWire-FG, and excels in CS and CO on MLB, highlighting its superiority in effectiveness. Furthermore, our method achieves only 40\% of Chain-of-Table's time and cost, showing its advantage in efficiency. In summary, Tree-of-Text opens a new path for prompt-based table-to-text generation in the sports domain.

\section*{Limitations}

% 我們的論文目前有一些 Limitations：
% - 不同的資料集需要不同的 configuration and prompt，這些目前都需要人工調整，缺乏一個自動化的方法。
% - Tree-of-Text 的 time and cost 雖然比 Tree-of-Thought 和 Chain-of-Table 低，但還是比 Few-shot 和 Chain-of-Thought 高。
% - Prompt-based 和 Model-based 的方法相比，雖然不需要大量的訓練資料，但使用 LLM 的成本還是相對較高，且資料的隱私性較低。

% \begin{itemize}
%     \item Different datasets require different configurations and prompts, which currently still need to be adjusted manually. There is no automated method for optimizing these settings.
%     \item Although Tree-of-Text achieves lower time and cost compared to Tree-of-Thought and Chain-of-Table, it is still higher than Few-shot and Chain-of-Thought.
%     \item Compared to model-based methods, prompt-based approaches do not require large-scale training data. However, the cost of using LLMs remains relatively high, and data privacy is a concern due to reliance on external models.
% \end{itemize}

While Tree-of-Text demonstrates strong performance, our approach requires manually tuning configurations and prompts for the corresponding dataset.
Therefore, one interesting research direction could be to explore automatic selection for configurations and prompts.

On the other hand, Tree-of-Text achieves better time and cost efficiency compared to Tree-of-Thought and Chain-of-Table, yet we leave the efficiency aspect as future work, as our proposed approach still requires more time and cost than Few-shot and Chain-of-Thought. Among possible solutions, using multi-threaded parallel processing might be a potential method.

Finally, Tree-of-Text can currently only access the internal information contained in the table and cannot obtain external information (e.g., player rankings, historical match records). Thus, designing new operations to call web crawlers for retrieving external information is also a promising direction for future research.

\section*{Ethical Considerations}

% Tree-of-Text 有以下兩個 Potential Risks。
% 第一，雖然 Tree-of-Text 相較於 model-based baselines 不需要大量的訓練資料，但使用外部的 LLM 還是會有資料隱私的疑慮，特別是在機密的資料上。
% 第二，儘管 Tree-of-Text 相較於 prompt-based baselines 有較高的資料保真度，但還是有可能會發生幻覺的問題，產生錯誤或不存在的資訊，導致讀者被這些假資訊誤導。

Firstly, although Tree-of-Text does not require large training datasets compared to model-based methods, using external LLMs raises concerns about data privacy, especially for sensitive information.  

Secondly, while Tree-of-Text achieves higher data fidelity compared to other prompt-based baselines, hallucination issues may still occur, potentially generating incorrect or non-existent information that could mislead readers.

\section*{Acknowledgments}

% In this work, 我們使用了 Copilot 來自動完成部份基礎的程式碼，我們也使用 ChatGPT 來修正文法、翻譯語言和尋找文獻。All uses were assessed to meet the ACL Rolling Review's AI Writing/Coding Assistance Policy.

In this work, we used Copilot to automatically complete some basic code and utilized ChatGPT for grammar corrections, language translation, and literature searches. All uses were reviewed to ensure compliance with the ACL Rolling Review's AI Writing/Coding Assistance Policy.

The ShuttleSet22 dataset is publicly available under the Creative Commons Attribution 4.0 International (CC BY 4.0) license. The RotoWire-FG dataset is released under the MIT License. The MLB dataset is licensed for non-commercial research purposes only. All datasets and associated artifacts are utilized exclusively for table-to-text generation research, and no commercial use is intended or permitted. During the data collection process, we manually verified that all reports do not contain any personally identifying information or offensive content.

\bibliography{custom}

\appendix

\section{Example Prompt}

\subsection{Example Prompt for Content Planning}
\label{sec:content_planning_prompt}

% Figure \ref{fig:content_planning_prompt} 是 Content Planning 在 ShuttleSet+ dataset 的範例 prompt。其中，{TABLE_DESCRIPTION} 是表格中每個欄位的說明文字，{OPERATION_DESCRIPTION} 則是每個 operation 的說明文字，{TABLES} 就是輸入的表格，{OPERATION_HISTORY} 是父節點使用過的 operation，{OPERATION_POOL} 則是尚未使用的 operation。最後 LLM 會根據以上的 prompt 輸出 operations \& arguments。

Figure \ref{fig:content_planning_prompt} shows an example prompt for Content Planning on the ShuttleSet+ dataset. 
In this prompt, \texttt{\{TABLE\_DESCRIPTION\}} provides descriptions for each column in the table, \texttt{\{OPERATION\_DESCRIPTION\}} explains each available operation, \texttt{\{TABLES\}} represents the input tables, \texttt{\{OPERATION\_HISTORY\}} lists the operations previously used by the parent node, and \texttt{\{OPERATION\_POOL\}} indicates the remaining unused operations. 
Then, the LLM outputs the \texttt{Operations \& Arguments} based on the input prompt.

\begin{figure*}[ht]
\begin{lstlisting}
System:

You are a content planner for the badminton game report.

Please select candidate Operations and corresponding Arguments from the Operation Pool based on the input Tables and Operation History. These candidate Operations will be the next Operation in the Operation History.

# Requirements

1. Strictly adhere to the requirements.
2. The output must be in English.
3. The output must be based on the input data; do not hallucinate.
4. The table format is {TABLE_FORMAT}.
5. The length of Operation History must be less than or equal to {MAX_DEPTH}.
6. The number of Operations must be less than or equal to {MAX_DEGREE}.
7. Only select Operations from the Operation Pool.
8. Arguments must match the format required by the corresponding Operations.
9. Operations & Arguments must follow this format: [operation_1(argument_1, ...), operation_2(argument_2, ...), operation_3(argument_3, ...), ...]
10. Only output Operations & Arguments!
11. The number of tokens in the Operations & Arguments must be within {PLANNING_TOKENS}.

# Table Description

{TABLE_DESCRIPTION}

# Operation Description

{OPERATION_DESCRIPTION}

User:

# Test

## Tables

{TABLES}

## Operation History

{OPERATION_HISTORY}

## Operation Pool

{OPERATION_POOL}

## Operations & Arguments


\end{lstlisting}
\vspace{-3.5mm}
\caption{Prompt for Content Planning}
\label{fig:content_planning_prompt}
\end{figure*}

\subsection{Example Prompt for \texttt{write()} operation}
\label{sec:write_prompt}

% Figure \ref{fig:write_prompt} 是 \texttt{write()} operation 在 ShuttleSet+ dataset 的範例 prompt。其中，{TABLE_DESCRIPTION} 是表格中每個欄位的說明文字，{TABLES} 就是輸入的表格。然後 LLM 會根據以上的 prompt 生成 Report。

Figure \ref{fig:write_prompt} shows an example prompt for the \texttt{write()} operation on the ShuttleSet+ dataset. 
In this prompt, \texttt{\{TABLE\_DESCRIPTION\}} provides descriptions for each column in the table, and \texttt{\{TABLES\}} represents the input tables. 
Then, the LLM generates the \texttt{Report} based on the input prompt.

\begin{figure*}[ht]
\begin{lstlisting}
System:

You are a content writer for the badminton game report.

Please write the Report based on the input Table.

# Requirements

1. Strictly adhere to the requirements.
2. The output must be in English.
3. The output must be based on the input data; do not hallucinate.
4. The Table format is {TABLE_FORMAT}.
5. The Report can only describe the content included in the Tables and cannot describe anything not included in the Tables.
6. The Report must consist of only one paragraph.
7. The number of tokens in the Report must be within {WRITE_TOKENS}.

# Table Description

{TABLE_DESCRIPTION}

User:

# Test

## Tables

{TABLES}

## Report


\end{lstlisting}
\vspace{-3.5mm}
\caption{Prompt for \texttt{write()} operation}
\label{fig:write_prompt}
\end{figure*}

\subsection{Example Prompt for Content Generating}
\label{sec:content_generating_prompt}

% Figure \ref{fig:content_generating_prompt} 是 Content Generating 在 ShuttleSet+ dataset 的範例 prompt。其中，{REPORTS} 是多篇報導。然後 LLM 會根據以上的 prompt merge and rewrite 一篇 \texttt{New Report}。

Figure \ref{fig:content_generating_prompt} shows an example prompt for Content Generating on the ShuttleSet+ dataset.
In this prompt, \texttt{\{REPORTS\}} represents multiple reports. 
Then, the LLM merges and rewrites the reports into a \texttt{New Report} based on the input prompt.

\begin{figure*}[ht]
\begin{lstlisting}
System:

You are a content generator for the badminton game report.

Please merge and rewrite a New Report based on the input Reports.

# Requirements

1. Strictly adhere to the requirements.
2. The output must be in English.
3. The output must be based on the input data; do not hallucinate.
4. The New Report must include all the content from the input Reports; do not omit any information.
5. The New Report must follow the order of the input Reports.
6. The number of tokens in the New Report must be within {GENERATING_TOKENS}.

User:

# Test

## Reports

{REPORTS}

## New Report


\end{lstlisting}
\vspace{-3.5mm}
\caption{Prompt for Content Generating}
\label{fig:content_generating_prompt}
\end{figure*}

\subsection{Example Prompt for the LLM-based IE model}
\label{sec:LLM_IE_model_prompt}

% Figure \ref{fig:LLM_IE_model_prompt} 是 LLM-based IE model 在 ShuttleSet+ dataset 的範例 prompt。其中，{TABLE_DESCRIPTION} 是表格中每個欄位的說明文字，{REPORT} 是一場比賽的報導，{TABLE_RELATION} 是這場比賽的表格中的所有 relation。然後，LLM 會根據以上的從 report 中擷取出 relation。

Figure \ref{fig:LLM_IE_model_prompt} shows an example prompt for the LLM-based IE model on the ShuttleSet+ dataset. 
In this prompt, \texttt{\{TABLE\_DESCRIPTION\}} provides descriptions for each column in the table, \texttt{\{REPORT\}} is a report of a match, and \texttt{\{TABLE\_RELATION\}} lists all relations from the match’s tables. 
The LLM then extracts relations from the report based on the input prompt.

\begin{figure*}[ht]
\begin{lstlisting}
System:

You are a relation extractor for the badminton game report.

Please extract the Report Relation contained in the Report from the Table Relation.

There is an Example that you can refer to.

# Requirements

1. Strictly adhere to the requirements.
2. The output must be in English.
3. The output must be based on the input data; do not hallucinate.
4. Please do not output any Report Relation that is not included in the Report.
5. Please do not output any Report Relation that is not included in the Table Relation.
6. The Report Relation must contain all the relations from the input Report; do not omit any relation.
7. The Report Relation must follow the order in the input Report.
8. The Report Relation must follow the format: [(table|column|value), (table|column|value), ...]

# Table Description

{TABLE_DESCRIPTION}

User:

# Test

## Report

{REPORT}

## Table Relation

{TABLE_RELATION}

## Report Relation

\end{lstlisting}
\vspace{-3.5mm}
\caption{Prompt for LLM-based IE model}
\label{fig:LLM_IE_model_prompt}
\end{figure*}

\section{Data Preprocessing}

\subsection{Data Preprocessing for ShuttleSet+}
\label{sec:data_preprocessing_for_shuttleset+}

% ShuttleSet22 是 stroke-level 的資料，但生成文字實際上並不需要用到如此細節的資訊，因此我們只保留了每個回合最後一球的行。此外，因為 ShuttleSet22 原本包含太多和文字不相關的欄位，所以我們只保留了九個最重要的欄位，並將部份欄位重新命名和重新排序，讓表格更加清楚簡潔。再來，因為 ball_type、win_reason、lose_reason 欄位的值是中文，我們將其翻譯成對應的英文。最後，我們將每場比賽整體的資訊加上每個 set 的分數，並獨立存成一個 match 的 CSV 表格。

ShuttleSet22 is a stroke-level dataset; however, generating textual descriptions does not require such detailed information. 
Therefore, we retain only the final stroke of each rally. 
To streamline the dataset, we selected the nine most essential columns, renaming and reordering to improve clarity while removing unrelated fields. 
Additionally, the values in the \texttt{ball\_type}, \texttt{win\_reason}, and \texttt{lose\_reason} columns were originally in Chinese, so we translated them into English. 
Lastly, we reorder the table columns according to the order specified in the table description of ShuttleSet+.

\subsection{Data Preprocessing for RotoWire-FG}
\label{sec: data_preprocessing_for_rotowire}

% 為了讓輸入符合 Tree-of-Text 的格式，我們先對 RotoWire-FG dataset 做預處理。首先，我們將原本的資料從 json 格式轉換成多個 csv 表格，分別是 game、home_line、vis_line、box_score，其中 game 表示整場比賽的資訊，home_line 表示主場的 line-score，vis_line 表示客場的 line-score，box_score 則表示每位球員在比賽中的分數。然後將表格欄位依照 Table Description 的順序重新排序。

To ensure the input format is aligned, we first preprocess the RotoWire-FG dataset. 
Initially, we convert the original data from JSON format into multiple CSV tables: \texttt{game}, \texttt{home\_line}, \texttt{vis\_line}, and \texttt{box\_score}. 
Specifically, \texttt{game} contains overall game information, \texttt{home\_line} represents the line score of the home team, \texttt{vis\_line} represents the line score of the visiting team, and \texttt{box\_score} records individual player statistics. 
Finally, we reorder the table columns according to the sequence specified in the table description of RotoWire-FG.

\subsection{Data Preprocessing for MLB}
\label{sec:data_preprocessing_for_mlb}

% 為了讓輸入符合 Tree-of-Text 的格式，我們也對 MLB dataset 做預處理。和 RotoWire-FG 類似，我們先將原本的資料從 json 格式轉換成多個 csv 表格，分別是 game, home_line, visiting_line, box_score, play_by_play，其中 game 表示整場比賽的資訊，home_line 表示主場的 line-score，vis_line 表示客場的 line-score，box_score 則表示每位球員在比賽中的分數，play_by_play 則表示每一次打擊的分數。然後將表格欄位依照 Table Description 的順序重新排序。最後，因為 box_score 有許多行都是 N/A，所以我們將這些冗餘的行移除。

To ensure the input format is aligned, we also preprocess the MLB dataset. Similar to RotoWire-FG, we first convert the original data from JSON format into multiple CSV tables: \texttt{game}, \texttt{home\_line}, \texttt{vis\_line}, \texttt{box\_score}, and \texttt{play\_by\_play}. 
Specifically, \texttt{game} contains overall game information, \texttt{home\_line} represents the line score of the home team, \texttt{vis\_line} represents the line score of the visiting team, \texttt{box\_score} records individual player statistics, and \texttt{play\_by\_play} details the scoring of each at-bat. 
Next, since the \texttt{box\_score} table contains many rows with \texttt{N/A} values, we remove these redundant rows to streamline the dataset.
Eventually, we reorder the table columns according to the sequence specified in the table description of MLB. 

\section{LLM-based IE model}
\label{sec:LLM_IE_model}

% 我們並沒有現成的 IE model for ShuttleSet+，訓練資料也不足，因此，我們改成使用 LLM 當作 IE model，從文字中擷取出 relation。為了證明 LLM-based IE model 的可信度，我們人工標記了幾筆 relation 並和 LLM 擷取的 relation 計算衡量指標，衡量結果圖表 \ref{tab:LLM_IE_model}。

Since no existing IE model is available for ShuttleSet+ and the training data is insufficient, we use an LLM as the IE model to extract information from the text. 
To validate the reliability of the LLM-based IE model, we manually annotated a set of information and compared it with that extracted by the LLM.
The evaluation results are presented in Table \ref{tab:LLM_IE_model}.

% 我們比較了 Zero-shot 和 One-shot 兩種 Prompt，實驗結果發現 One-shot 的表現較好，在所有指標上都大於 60％，因此後續都會用 One-shot 當作 LLM-based IE model 的 Prompt。推測原因是因為多給了一個範例，讓 LLM 可以參考這個範例，擷取出更接近人工範例的 relation。One-shot 的 prompt 如 Appendix \ref{sec:LLM_IE_model_prompt} 所示.

We compared three prompting methods for the LLM-based IE model: Zero-shot, One-shot, and Few-shot. 
Experimental results show that Few-shot performs better, achieving over 70\% across all metrics. 
Therefore, we used Few-shot prompting for all subsequent experiments. 
We hypothesize that providing more examples allows the LLM to reference them, leading to information extraction that more closely aligns with human annotations. 
The prompt for the LLM-based IE model is provided in Appendix~\ref{sec:LLM_IE_model_prompt}.

\begin{table}[]
\resizebox{\columnwidth}{!}{%
\begin{tabular}{@{}ccccccc@{}}
\toprule
\textbf{Prompt} & \textbf{RG \#}   & \textbf{RG P\%} & \textbf{CS P\%} & \textbf{CS R\%} & \textbf{CS F\%} & \textbf{CO DLD\%} \\ \midrule
Zero-shot       & \textbf{14.0000} & \textbf{100.00} & 70.56           & \textbf{76.57}  & {\ul 71.51}     & 26.80             \\
One-shot        & {\ul 12.3333}    & \textbf{100.00} & {\ul 75.35}     & {\ul 70.46}     & 70.71           & {\ul 38.24}       \\
\rowcolor[HTML]{FFFFC7} 
Few-shot        & 10.3333          & \textbf{100.00} & \textbf{93.89}  & \textbf{76.57}  & \textbf{83.86}  & \textbf{71.01}    \\ \bottomrule
\end{tabular}%
}
\caption{The evaluation results of the LLM-based IE model with different prompts, where the best scores are highlighted in \textbf{bold}, the second-best scores are \underline{underlined}, and the best configuration is marked with a yellow background.}
\label{tab:LLM_IE_model}
\end{table}

\section{Ablation Study}

\subsection{The Effects of Large Language Models}

To validate the generalizability of Tree-of-Text and examine the impact of model size on performance, we conducted additional experiments on ShuttleSet+ using open-source LLMs of different sizes (e.g., llama3.1-8b, llama3.1-70b, llama3.1-405b, and deepseek-v3) and closed-source LLMs (e.g., gpt-4o-mini, gpt-4o) as the backbone LLMs.
As open-source models are freely available for download and use, we treat their cost as zero.
The results are presented in Table \ref{tab:LLMs}.

% 結果顯示，模型越大表現越好，這是因為越大的模型越可以遵守 prompt 的指示並輸出預期的結果，但相對的時間和成本就越高。
% 此外，llama3.1-405b 的表現只比 gpt-4o-mini 差一點點，驗證我們的方法在 open-source LLMs LLMs 的泛用性。
% 然而，gpt-4o 並沒有勝過 gpt-4o-mini，表示 gpt-4o-mini 已經做的夠好了，因為時間和成本考量，我們最後還是選擇 gpt-4o-mini as the backbone LLM.

% The results show that as the model size increases, performance improves, but conversely, both time and cost also increase. 
% This is because larger models are better able to adhere to the prompts and produce the expected results.
% 此外，deepseek-v3 的表現甚至比 gpt-4o-mini 還更好，可見 Mixture-of-Experts (MoE) 模型的優勢，也 validating the generalizability of our method on open-source LLMs.
% However, gpt-4o did not outperform gpt-4o-mini, suggesting that gpt-4o-mini already performs sufficiently well on this task. 
% Considering both time and cost factors, we ultimately chose gpt-4o-mini as the backbone LLM.

The results indicate that increasing model size leads to improved performance; however, this comes at the expense of higher computational time and cost. Larger models are better able to follow prompts and generate the expected outputs. Furthermore, deepseek-v3 even outperforms gpt-4o-mini, highlighting the advantages of Mixture-of-Experts (MoE) models and validating the generalizability of our approach on open-source LLMs. Interestingly, gpt-4o does not surpass gpt-4o-mini, suggesting that gpt-4o-mini is already sufficiently capable for this task. Considering both time and cost, we ultimately select gpt-4o-mini as the backbone LLM.

\subsection{The Analysis of Max Depth \& Max Degree}

% 為了找到最適合 Tree-of-Text 的 max depth and max degree，我們進行了以下實驗：首先對照組的 max depth and max degree 都是 5，其中一組實驗的 max depth 改成 3，另外一組實驗的 max degree 改成 3，而最後一組實驗的 max depth and max degree 都設為 3。實驗結果如表 \ref{tab:max_depth_max_degree}.

To determine the optimal maximum depth and maximum degree for Tree-of-Text, we conducted the following experiments. 
The baseline setting uses a max depth and max degree of 5. 
In one experiment, we reduced the max depth to 3 while keeping the max degree at 5. 
In another, we set the max degree to 3 while maintaining the max depth at 5. 
Finally, we tested a configuration where both the max depth and max degree were set to 3. 
The experimental results on ShuttleSet+ are presented in Table~\ref{tab:max_depth_max_degree}.

% 從實驗中可以觀察到，max depth and max degree 都是 5 的表現是最好的，但 time and cost 也比較高。當我們把 max depth 改成 3 時，相較於把 max degree 改成 3，表現下降的幅度是更明顯的，也就是說 max depth 的影響力比 max degree 還要大，這也是可以理解的，因為 max depth 是控制文字的細節程度，而 max degree 則是控制文字的豐富程度。

From the experimental results, we observe that setting both max depth and max degree to 5 yields the best performance; however, it also results in higher time and cost. 
When reducing the max depth to 3 while keeping the max degree at 5, the performance drops more significantly compared to reducing the max degree to 3 while keeping the max depth at 5. 
This suggests that max depth has a greater impact on the generated text than max degree. 
This finding is intuitive, as max depth controls the level of detail in the text, whereas max degree influences its richness.

% 最後，將 max depth and max degree 都設為 3 的結果當然是最差的，但值得注意的是，它的 time and cost 也是最低的。因此，我們可以根據文字來調整 max depth and max degree。當我們想要生成比較詳細的文字時，max depth and max degree 就可以調高一點，但成本就會比較高。而當我們想要生成比較簡略的文字時，max depth and max degree 就可以調低一點，但成本就會比較低。

Finally, setting both max depth and max degree to 3 yields the worst performance, as expected. 
However, it is worth noting that this setting also results in the lowest time and cost. 
This suggests that max depth and max degree can be adjusted based on the desired level of detail in the generated text. 
If more detailed text is required, increasing max depth and max degree improves performance at the expense of higher computational cost. 
Conversely, for more general text, reducing the max depth and max degree lowers both the level of detail and cost.

\subsection{The Influences of Operation Pool}

% 為了證明每個 operation 都是有作用的，我們進行了以下實驗：首先，對照組是 operation pool 中包含全部的 operations。然後我們依序從 operation pool 中移除其中一個 operation 來進行實驗。實驗結果請看表 \ref{tab:operation_pool}.

To demonstrate the significance of each operation, we performed the following experiments. 
The baseline configuration includes all operations in the operation pool. 
Then, in each experiment, we systematically removed one operation from the operation pool and evaluated the impact on performance. 

The experimental results on ShuttleSet+ in Table~\ref{tab:operation_pool} show that removing \texttt{select\_table()}, \texttt{select\_row()}, and \texttt{select\_col()} leads to a significant performance drop, highlighting their importance.
Without these operations, the LLM processes the entire table to generate text, leading to increased time and cost. 
In contrast, removing \texttt{count()}, \texttt{sort()}, and \texttt{filter()} has a less pronounced effect, suggesting that they are relatively less critical. 
However, without these operations, it becomes impossible to compute more detailed information, resulting in a lower RG \# but a higher RG P\%.
Overall, maintaining all operations provides the most balanced performance, demonstrating greater robustness.

\begin{figure*}[t]
  \includegraphics[width=\linewidth]{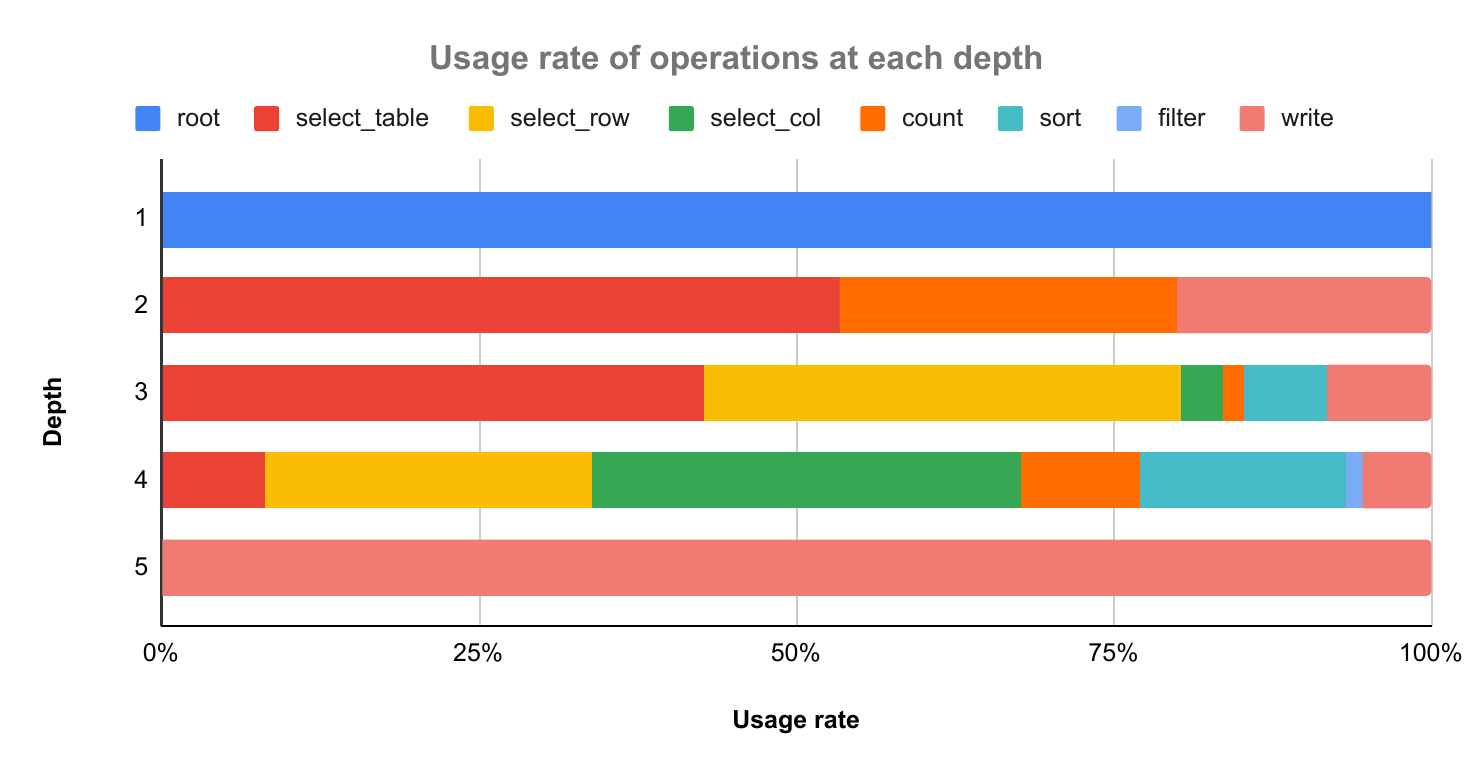}
  \caption{Usage rate of operations at each level, where the horizontal axis indicates the usage rate, the vertical axis denotes the depth of the tree, and different colors represent different operations.}
  \label{fig:usage_rate_of_operations}
\end{figure*}

% 此外，我們計算了 Usage rate of operations at each level 並繪製成 Figure~\ref{usage_rate_of_operations}。從圖中我們可以看出，\texttt{select\_table()}, \texttt{select\_row()}, and \texttt{select\_col()} 主要都在深度低，而 \texttt{count()}, \texttt{sort()}, and \texttt{filter()} 則都在深度高，可見 LLM 知道要先篩選出比較相關的表格，再去統計出詳細資訊，也驗證了 \texttt{select\_table()}, \texttt{select\_row()}, and \texttt{select\_col()} 比 \texttt{count()}, \texttt{sort()}, and \texttt{filter()} 的影響力更大。

In addition, we calculate the usage rate of operations at each level and visualize the results in Figure~\ref{fig:usage_rate_of_operations}. From the figure, we observe that \texttt{select\_table()}, \texttt{select\_row()}, and \texttt{select\_col()} are predominantly used at lower depths, while \texttt{count()}, \texttt{sort()}, and \texttt{filter()} tend to appear at higher depths. This indicates that the LLM first selects the most relevant tables before performing more detailed computations, validating that \texttt{select\_table()}, \texttt{select\_row()}, and \texttt{select\_col()} have a greater impact than \texttt{count()}, \texttt{sort()}, and \texttt{filter()}.

\subsection{The Impacts of Table Formats}

% 我們也比較了不同表格格式對 Tree-of-Text 表現的影響，我們比較了四種常見的表格格式 CSV (Comma-Separated Values), PIPE, Markdown, HTML (HyperText Markup Language)。CSV uses commas to separate values, ideal for spreadsheets. PIPE uses | as a delimiter, often in command-line outputs. Markdown tables use pipes (|) for columns and hyphens (-) for headers, common in documentation. HTML tables use <table>, <tr>, and <td> tags for structured web display.

We also analyzed the impact of different table formats on Tree-of-Text's performance by comparing four commonly used formats: CSV (Comma-Separated Values), PIPE, Markdown, and HTML (HyperText Markup Language). 
CSV separates values with commas, making it ideal for spreadsheets, while PIPE uses the \texttt{|} symbol as a delimiter, commonly found in command-line outputs. 
Markdown tables employ pipes (\texttt{|}) to define columns and hyphens (\texttt{-}) for headers, frequently used in documentation. 
HTML tables utilize \texttt{<table>}, \texttt{<tr>}, and \texttt{<td>} tags to structure tabular data for the website.

% 從表格 \ref{tab:table_format} 的實驗結果中可以看出，CSV 的表現是最好的，而 PIPE 和 HTML 表現雖然差不多，但 time and cost 明顯較高，這是因為他們需要更多符號表示表格，所以 input context 比較長。至於 Markdown 的表現是最差的，這可能是因為 LLM 對於這類型的訓練資料比較少，導致表格理解能力較低。基於以上實驗結果，我們最後採用 CSV 當作後續實驗的 table format。

The experimental results on ShuttleSet+ in Table~\ref{tab:table_format} show that CSV achieves the best performance. 
While PIPE and HTML perform similarly, they have significantly higher time and cost due to requiring more symbols to represent the table, resulting in a longer input context. 
Markdown performs the worst, likely because LLMs have been pre-trained on fewer examples of this format, leading to weaker table comprehension.

\begin{table*}
\small
\centering
\begin{tabular}{@{}ccccccccc@{}}
\toprule
\textbf{LLMs}                                            & \textbf{RG \#} & \textbf{RG P\%} & \textbf{CS P\%} & \textbf{CS R\%} & \textbf{CS F\%} & \textbf{CO DLD\%} & \textbf{Time}  & \textbf{Cost} \\ \midrule
llama3.1-8b                                              & 7.44           & 49.21           & 45.94           & 43.14           & 43.95           & 42.48             & \textbf{10.91} & \textbf{0.00} \\
llama3.1-70b                                             & 11.33          & 86.57           & 64.53           & 68.54           & 62.50           & 59.18             & 33.12          & \textbf{0.00} \\
llama3.1-405b                                            & 15.56          & 96.17           & 93.89           & 91.98           & 92.77           & 91.98             & 57.16          & \textbf{0.00} \\
deepseek-v3                                              & \textbf{16.25} & \textbf{98.33}  & \textbf{94.01}  & \textbf{94.12}  & \textbf{94.06}  & 92.64             & 89.25          & \textbf{0.00} \\ \midrule
\rowcolor[HTML]{FFFFC7} gpt-4o-mini & {\ul 15.78}    & {\ul 98.04}     & {\ul 93.94}     & {\ul 93.94}     & {\ul 93.94}     & \textbf{93.94}    & {\ul 29.04}    & {\ul 5.71}    \\
gpt-4o                                                   & {\ul 15.78}    & {\ul 98.04}     & 93.29           & 93.29           & 93.29           & {\ul 93.29}       & 33.73          & 54.88         \\ \bottomrule
\end{tabular}
\caption{The experimental results for different LLMs on ShuttleSet+, where the best scores are highlighted in \textbf{bold}, the second-best scores are \underline{underlined}, and the best configuration is marked with a yellow background.}
\label{tab:LLMs}
\end{table*}

% \begin{table}[]
% \resizebox{\columnwidth}{!}{%
% \begin{tabular}{@{}cccccccccc@{}}
% \toprule
% \textbf{Max Depth} & \textbf{Max Degree} & \textbf{RG \#} & \textbf{RG P\%} & \textbf{CS P\%} & \textbf{CS R\%} & \textbf{CS F\%} & \textbf{CO DLD\%} & \textbf{Time}  & \textbf{Cost} \\ \midrule
% \rowcolor[HTML]{FFFFC7} 
% 5                  & 5                   & \textbf{15.78} & \textbf{98.04}  & \textbf{93.94}  & \textbf{93.94}  & \textbf{93.94}  & \textbf{93.94}    & {\ul 29.04}    & \textbf{5.71} \\
% 3                  & 5                   & {\ul 15.67}    & 95.99           & 91.42           & {\ul 92.72}     & 92.03           & 91.42             & 16.60          & 2.37          \\
% 5                  & 3                   & {\ul 15.67}    & {\ul 97.21}     & {\ul 92.46}     & 92.46           & {\ul 92.46}     & {\ul 92.46}       & \textbf{29.48} & {\ul 4.90}    \\
% 3                  & 3                   & 13.89          & 88.18           & 83.43           & 82.00           & 82.56           & 82.00             & 11.79          & 1.82          \\ \bottomrule
% \end{tabular}%
% }
% \caption{The experimental results for comparing different max depth and max degree on ShuttleSet+, where the highest scores are highlighted in \textbf{bold}, the second-highest scores are \underline{underlined}, and the best configuration is marked with a yellow background.}
% \label{tab:max_depth_max_degree}
% \end{table}

\begin{table*}
\small
\centering
\begin{tabular}{@{}cccccccccc@{}}
\toprule
\textbf{Max Depth} & \textbf{Max Degree} & \textbf{RG \#} & \textbf{RG P\%} & \textbf{CS P\%} & \textbf{CS R\%} & \textbf{CS F\%} & \textbf{CO DLD\%} & \textbf{Time}  & \textbf{Cost} \\ \midrule
\rowcolor[HTML]{FFFFC7} 
5                  & 5                   & \textbf{15.78} & \textbf{98.04}  & \textbf{93.94}  & \textbf{93.94}  & \textbf{93.94}  & \textbf{93.94}    & 29.04    & 5.71 \\
3                  & 5                   & {\ul 15.67}    & 95.99           & 91.42           & {\ul 92.72}     & 92.03           & 91.42             & {\ul 16.60}          & {\ul 2.37}          \\
5                  & 3                   & {\ul 15.67}    & {\ul 97.21}     & {\ul 92.46}     & 92.46           & {\ul 92.46}     & {\ul 92.46}       & 29.48 & 4.90    \\
3                  & 3                   & 13.89          & 88.18           & 83.43           & 82.00           & 82.56           & 82.00             & \textbf{11.79}          & \textbf{1.82}          \\ \bottomrule
\end{tabular}%
\caption{The experimental results for comparing different max depth and max degree on ShuttleSet+, where the best scores are highlighted in \textbf{bold}, the second-best scores are \underline{underlined}, and the best configuration is marked with a yellow background.}
\label{tab:max_depth_max_degree}
\end{table*}

\begin{table*}
\small
\centering
\begin{tabular}{@{}ccccccccc@{}}
\toprule
\textbf{Operation Pool} & \textbf{RG \#} & \textbf{RG P\%} & \textbf{CS P\%} & \textbf{CS R\%} & \textbf{CS F\%} & \textbf{CO DLD\%} & \textbf{Time}  & \textbf{Cost} \\ \midrule
\rowcolor[HTML]{FFFFC7} 
All operations          & \textbf{15.78} & {\ul 98.04}     & \textbf{93.94}  & 93.94           & \textbf{93.94}  & \textbf{93.94}    & {\ul 29.04}          & 5.71          \\
w/o \texttt{select\_table()}     & {\ul 15.44}    & \textbf{98.69}  & 82.57           & 92.94           & 85.57           & 82.57             & 44.45          & 6.11          \\
w/o \texttt{select\_row()}       & 15.33          & {\ul 98.04}     & 84.53           & {\ul 94.90}     & 87.53           & 84.53             & 48.00    & 6.80    \\
w/o \texttt{select\_col()}       & 15.11          & \textbf{98.69}  & {\ul 85.19}     & 93.33           & 86.95           & 82.96             & 49.80 & 7.49 \\
w/o \texttt{count()}             & {\ul 15.44}    & \textbf{98.69}  & 82.57           & 92.94           & 85.57           & 82.57             & \textbf{25.30}          & \textbf{4.20}          \\
w/o \texttt{sort()}              & {\ul 15.44}    & \textbf{98.69}  & {\ul 85.19}     & \textbf{95.56}  & {\ul 88.18}     & {\ul 85.19}       & 36.64          & 5.64          \\
w/o \texttt{filter()}            & {\ul 15.44}    & \textbf{98.69}  & 82.57           & 92.94           & 85.57           & 82.57             & 33.34          & {\ul 5.53}          \\ \bottomrule
\end{tabular}%
\caption{The experimental results for comparing different operation pools on ShuttleSet+, where the best scores are highlighted in \textbf{bold}, the second-best scores are \underline{underlined}, and the best configuration is marked with a yellow background.}
\label{tab:operation_pool}
\end{table*}

% \begin{table}[]
% % \small
% \resizebox{\columnwidth}{!}{%
% \begin{tabular}{@{}ccccccccc@{}}
% \toprule
% \textbf{Table Format} & \textbf{RG \#} & \textbf{RG P\%} & \textbf{CS P\%} & \textbf{CS R\%} & \textbf{CS F\%} & \textbf{CO DLD\%} & \textbf{Time}   & \textbf{Cost}  \\ \midrule
% \rowcolor[HTML]{FFFFC7} 
% CSV                   & \textbf{15.78} & \textbf{98.04}  & \textbf{93.94}  & \textbf{93.94}  & \textbf{93.94}  & \textbf{93.94}    & 29.04           & 5.71           \\
% PIPE                  & \textbf{15.78} & \textbf{98.04}  & {\ul 93.29}     & {\ul 93.29}     & {\ul 93.29}     & {\ul 93.29}       & {\ul 78.53}     & 9.63           \\
% HTML                  & {\ul 15.67}    & {\ul 97.39}     & 92.64           & 92.64           & 92.64           & 92.64             & \textbf{104.99} & \textbf{19.26} \\
% Markdown              & 14.67          & 92.31           & 87.56           & 86.75           & 87.10           & 84.14             & 62.65           & {\ul 9.80}     \\ \bottomrule
% \end{tabular}%
% }
% \caption{The experimental results for comparing different table formats on ShuttleSet+, where the highest scores are highlighted in \textbf{bold}, the second-highest scores are \underline{underlined}, and the best configuration is marked with a yellow background.}
% \label{tab:table_format}
% \end{table}

\begin{table*}
\small
\centering
\begin{tabular}{@{}ccccccccc@{}}
\toprule
\textbf{Table Format} & \textbf{RG \#} & \textbf{RG P\%} & \textbf{CS P\%} & \textbf{CS R\%} & \textbf{CS F\%} & \textbf{CO DLD\%} & \textbf{Time}   & \textbf{Cost}  \\ \midrule
\rowcolor[HTML]{FFFFC7} 
CSV                   & \textbf{15.78} & \textbf{98.04}  & \textbf{93.94}  & \textbf{93.94}  & \textbf{93.94}  & \textbf{93.94}    & \textbf{29.04}           & \textbf{5.71}           \\
PIPE                  & \textbf{15.78} & \textbf{98.04}  & {\ul 93.29}     & {\ul 93.29}     & {\ul 93.29}     & {\ul 93.29}       & 78.53     & {\ul 9.63}           \\
HTML                  & {\ul 15.67}    & {\ul 97.39}     & 92.64           & 92.64           & 92.64           & 92.64             & 104.99 & 19.26 \\
Markdown              & 14.67          & 92.31           & 87.56           & 86.75           & 87.10           & 84.14             & {\ul 62.65}           & 9.80     \\ \bottomrule
\end{tabular}%
\caption{The experimental results for comparing different table formats on ShuttleSet+, where the best scores are highlighted in \textbf{bold}, the second-best scores are \underline{underlined}, and the best configuration is marked with a yellow background.}
\label{tab:table_format}
\end{table*}

% \section{Qualitative Result}
% \label{sec:qualitative_result}

\begin{figure*}[t]
  \includegraphics[width=\linewidth]{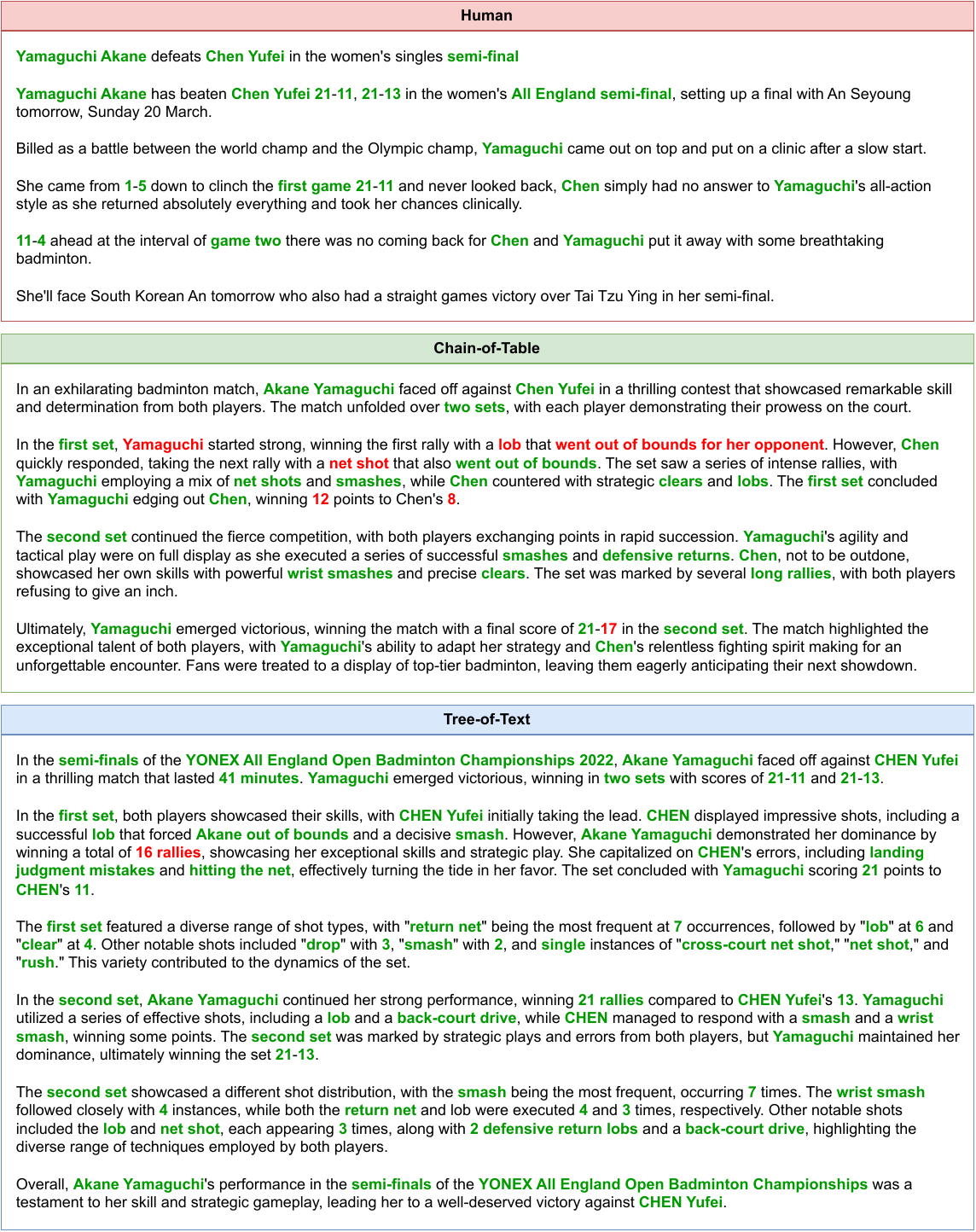}
  \caption{The qualitative results of Human, Chain-of-Table, and Tree-of-Text outputs.}
  \label{fig:qualitative_result}
\end{figure*}

% Figure~\ref{fig:qualitative_result} showcases the qualitative results of Human, Chain-of-Table, and Tree-of-Text outputs. 
% For ease of comparison, we mark the information in the text with \textbf{bold}: green indicates information included in the tables, while red indicates errors or information not found in the tables.  

% From the qualitative results, we observe that compared to Chain-of-Table, Tree-of-Text generates more comprehensive and detailed information (e.g., shot type frequencies) and produces more accurate outputs, with only one error compared to seven errors from Chain-of-Table. 
% This further validates that Tree-of-Text can generate text that meets the characteristics of high data fidelity and long-form output for sports game reports.

\begin{algorithm*}
% \small
% \footnotesize
% \scriptsize
% \tiny
\caption{Tree-of-Text}
\label{alg:overview}
\begin{algorithmic}[1]
\Require Tables $T$, Operation History $OH$, Operation Pool $OP$, Depth $D$, Max Depth $MAX\_DEPTH$, Max Degree $MAX\_DEGREE$
\Ensure Text $t$

\Function{Tree-of-Text}{$T, OH, OP, D$}

    \Comment{Depth must not exceed Max Depth}
    
    \If{$D \geq MAX\_DEPTH$} 
        \State $t \gets \Call{write}{T}$
        \State \Return $t$
    \EndIf

    \Comment{Content Planning}
    
    \State $OA \gets \Call{content\_planning}{T, OH, OP}$

    \Comment{Operation Execution}
    
    \State $t' \gets ()$
    \For{each $O_i ( A_i ) \mid O_i \in \text{OP}, i = 1, 2, \ldots, d$ in $OA$}
    
        \Comment{Degree must not exceed Max Degree}
        
        \If{$i \geq MAX\_DEGREE$}
            \State \textbf{break}
        \EndIf
        
        \If{$O_i = write()$}
            \State $t_i' \gets \Call{write}{T}$
        \Else
            \State $T_i' \gets O_i ( T, A_i )$
            \State $OH_i' \gets OH + O_i(A_i)$
            \State $OP_i' \gets OP - O_i()$
            \State $D_i' \gets D + 1$
            \State $t_i' \gets \Call{Tree-of-Report}{T_i', OH_i', OP_i', D_i'}$
        \EndIf
        \State $t' \gets t' + t_i'$
    \EndFor
    
    \Comment{Content Generating}
    
    \State $t \gets \Call{content\_generating}{t'}$
    
    \State \Return $t$
\EndFunction

\Statex

\State \textbf{Main Program}
\State $T \gets ( T^j \mid j = 1, 2, \ldots, n )$
\State $OH \gets ( op \mid op = \text{root()} )$
\State $OP \gets ( op \mid op \in \text{operations},\ op \neq \text{root()} )$
\State $D \gets 0$

\State $t \gets \Call{Tree-of-Text}{T, OH, OP, D}$

\end{algorithmic}
\end{algorithm*}

\end{document}